\documentclass[letterpaper, 10 pt, conference]{ieeeconf}  

\IEEEoverridecommandlockouts                              

\usepackage{url}
\usepackage{graphicx}
\usepackage{float}
\usepackage{url}   
 \usepackage{amsmath}

\title{\LARGE \bf
HumanMCP: A Human-Like Query Dataset for Evaluating MCP Tool Retrieval Performance
}

\author{
Shubh Laddha,
Lucas Changbencharoen,
Win Kuptivej,
Surya Shringla,
Archana Vaidheeswaran,
Yash Bhaskar}

\begin{document}

\maketitle
\thispagestyle{empty}
\pagestyle{empty}

\begin{abstract}

Model Context Protocol (MCP) servers contain a collection of thousands of open-source standardized tools, linking LLMs to external systems; however, existing datasets and benchmarks lack realistic, human-like user queries, remaining a critical gap in evaluating the tool usage and ecosystems of MCP servers. Existing datasets often do contain tool descriptions but fail to represent how different users portray their requests, leading to poor generalization and inflated reliability of certain benchmarks. This paper introduces the first large-scale MCP dataset featuring diverse, high-quality diverse user queries generated specifically to match 2800 tools across 308 MCP servers, developing on the MCP Zero dataset. Each tool is paired with multiple unique user personas that we have generated, to capture varying levels of user intent ranging from precise task requests, and ambiguous, exploratory commands, reflecting the complexity of real-world interaction patterns.

\end{abstract}

\section{INTRODUCTION}
Large language models (LLMs) have become central to modern AI, enabling reasoning, summarization, and tool use across a wide range of domains \cite{brown2020language, openai2023gpt4}. The Model Context Protocol (MCP) provides a unified way to connect LLMs to tools hosted on independently developed servers \cite{hou2025mcp}. MCP standardizes tool access through a JSON-RPC interface and exposes structured metadata describing each tool’s name, purpose, and inputs \cite{fei2025mcp, mcp_tools_docs}, allowing models to invoke tools directly from natural-language queries.

Despite this standardization, evaluating tool-using models remains difficult. Existing benchmarks largely focus on tool descriptions and task definitions but rarely capture the variability of real human communication \cite{gao2025mcp-radar, wang2025mcpbench, fan2025mcptoolbenchpp}. Real user queries differ widely in clarity, intent, and phrasing, whereas synthetic prompts tend to be uniform and idealized. As a result, current evaluations often inflate model performance and fail to reflect realistic tool-use scenarios.

To address this gap, we introduce a two-stage Generator Critic pipeline that produces realistic, semantically grounded user queries directly from tool metadata. The system generates diverse query styles by assigning five user personas ranging from precise to vague and from novice to expert. Using this method, we create a large dataset containing queries for approximately 2,800 tools across 308 MCP servers. The dataset is designed to reflect natural communication patterns rather than template-based prompts, enabling more accurate evaluation and development of MCP-based language-agent systems.

\section{RELATED WORKS}

The research on evaluating tool-using LLMs under the MCP has skyrocketed rapidly, leading to a range of benchmarks that address varying dimensions of tool-facilitated reasoning and performance. These works collectively lay the foundations for understanding model performance across varying MCP environments.

MCPToolBench++ \cite{fan2025mcptoolbenchpp} introduced one of the first large-scale benchmarks for tool-based LLM evaluations. It showcased automated pipelines for extracting tool metadata, generating structured queries whilst assessing functionality coverage across thousands of MCP servers. This work demonstrated the feasibility of large-scale, schema-based benchmarking in the MCP ecosystem.

Further developing on this, MCP-Bench  \cite{wang2025mcpbench} proposed a unified evaluation framework that assessed accuracy, completion efficiency, and cross-tool coordination. It formalized performance metrics for multi-step reasoning tasks, establishing a baseline for comparing LLMs under shared communication protocols.

MCP-AgentBench \cite{guo2025mcpagentbench} also stretched the benchmarking scope to agent-level reasoning by introducing categorized task types. Its structured taxonomy improved comprehensibility and comparability, allowing for the systematic measurement of agent performance across tool-mediated tasks.

Recently, TOUCAN \cite{xu2025toucan} contributed a large-scale dataset capturing over 1.5 million tool-agent interaction trajectories from real MCP environments. This focused on execution realism and behavioral data, providing insight into how agents interact dynamically across servers and tools.

Together, these efforts have helped progress and expand MCP benchmarking from a schema-based task evaluation to execution-level realism. Our present work continues this path by introducing a linguistically grounded dataset designed to evaluate user-agent interaction at a more naturalistic level. By focusing on human-like, persona-conditioned user queries, our study complements existing MCP benchmarks. It contributes to a more comprehensive understanding of language model performance in realistic, multi-tool environments.

\begin{figure*}[t]
    \vspace*{1em}
    \centering
    \framebox{
        \parbox{0.95\textwidth}{
            \centering
            \includegraphics[width=0.95\textwidth]{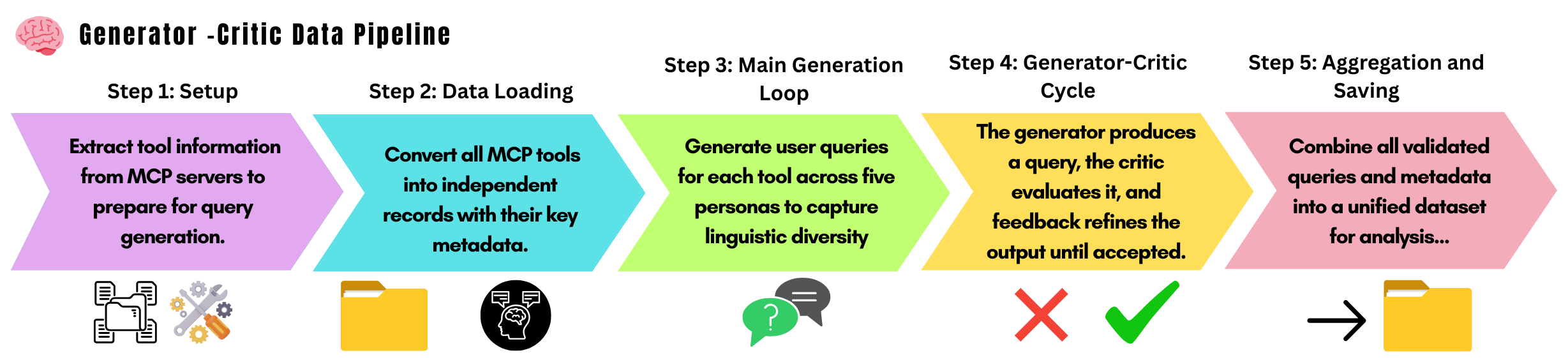}
        }
    }
    \caption{The Flowchart for our Generator–Critic Data Pipeline}
    \label{fig:placeholder}
\end{figure*}

\section{Methodology}

\subsection{Pipeline Overview}
We implemented a two-stage \textit{Generator-Critic} pipeline to synthesize realistic user queries from structured tool metadata. The system utilizes \texttt{gpt-4o-mini} as a cost-efficient generator and \texttt{gpt-4o} as a high-reasoning critic. The pipeline transforms static tool definitions into natural-language requests, employing a feedback loop with a maximum retry limit of two to balance computational efficiency with output quality.

\vspace{4pt}

\begin{table}[H]
\centering

\setlength{\tabcolsep}{6pt}      
\setlength{\arrayrulewidth}{0.7pt} 
\renewcommand{\arraystretch}{1.25} 

\caption{User Personas and Their Behavioral Descriptions}
\label{tab:personas}

\resizebox{\columnwidth}{!}{%
\begin{tabular}{|p{0.34\columnwidth}|p{0.62\columnwidth}|}
\hline
\textbf{Persona} & \textbf{Behavior Description} \\ \hline

\textbf{Problem-Oriented} &
Novices who describe issues in simple, non-technical terms. \\ \hline

\textbf{Goal-Oriented} &
Users who know what they want to achieve but not how to accomplish it. \\ \hline

\textbf{Category-Aware} &
Users aware of the general type/category of the tool but not its specific name. \\ \hline

\textbf{Function-Specific} &
Technically skilled users who describe precise functions and parameters without naming the tool. \\ \hline

\textbf{Tool-Explicit} &
Expert users who directly reference the exact tool name and may specify parameters explicitly. \\ \hline

\end{tabular}
}
\end{table}

\subsection{Persona-Based Query Generation}
To ensure the dataset captures the spectrum of real-world user behaviors, we defined five distinct personas ranging from novice to expert. For each tool, the generator produced queries corresponding to the specific intent and ambiguity levels defined in Table \ref{tab:personas}.

\subsection{The Generator-Critic Cycle}
The generation process follows an adversarial loop. First, the generator constructs a candidate query based on the tool's metadata and the assigned persona. The critic model immediately evaluates the candidate against three criteria: functional relevance, persona fidelity, and linguistic naturalness. It outputs a structured JSON verdict (\texttt{PASS} or \texttt{FAIL}) with reasoning. If a query fails, the critic's feedback guides a regeneration step. This self-correcting mechanism ensures that the final dataset remains semantically accurate while reflecting the specific intent of the target persona.

\section{Experiments}
We designed three experiments to validate the dataset's utility for benchmarking tool selection. These experiments assess baseline model performance, long-context scalability, and the dataset's applicability to Retrieval-Augmented Generation (RAG) pipelines.

\subsection{Experiment 1: Cross-Model Baseline}
This experiment establishes baseline performance across varying context sizes to verify the dataset's ability to differentiate model capabilities.

\vspace{0.5em}
\textbf{Models.} We evaluated three efficiency-optimized LLMs: \textbf{GPT-4o-mini} (OpenAI), \textbf{Gemini 2.0 Flash} (Google), and \textbf{Claude 3.5 Haiku} (Anthropic).

\vspace{0.5em}
\textbf{Configuration.} A fixed set of 500 queries was tested across three context configurations: 10, 50, and 100 tools. In each trial, the context contained one correct tool and $N-1$ distractors.

\vspace{0.5em}
\textbf{Protocol.} For each query, we dynamically sampled distractors, formatted the tool definitions into a standard prompt, and measured the \textbf{Top-1 Hit Rate} (exact match accuracy). We also calculated \textbf{Average Degradation} to quantify the performance drop as the context size increased from 10 to 100 tools.

\subsection{Experiment 2: Long-Context Scaling}
To evaluate the dataset's robustness at extreme scales, we stress-tested \textbf{Gemini 2.0 Flash}, chosen for its 1-million-token context window. Using the same protocol as Experiment 1, we expanded the search space to \textbf{500, 1,000, and 2,000 tools}. This experiment determines whether the dataset remains semantically distinct enough to support retrieval in massive, high-noise environments.

\subsection{Experiment 3: RAG Pipeline Evaluation}
This experiment assesses a realistic two-stage workflow where a lightweight retriever filters candidates before LLM selection.

\vspace{0.5em}
\textbf{Setup:} We utilized a fixed pool of 2,000 tools. To prevent data leakage, retrievers were calibrated on a separate training set of 2,800 query-tool pairs that were excluded from the evaluation pool.

\vspace{0.5em}
\textbf{Retrievers:} We compared three baselines:
\begin{enumerate}
    \item \textbf{TF-IDF:} Sparse vectorization with cosine similarity.
    \item \textbf{BM25:} Probabilistic ranking based on term frequency.
    \item \textbf{MiniLM (SentenceTransformer):} Dense retrieval using \texttt{all-MiniLM-L6-v2} embeddings.
\end{enumerate}

\vspace{0.5em}
\textbf{Pipeline:} 
\begin{enumerate}
    \item \textbf{Retrieval:} The classical model scores all 2,000 tools and returns the top 10 candidates.
    \item \textbf{Selection:} Gemini 2.0 Flash receives the query and the top 10 tool descriptions (without ranking scores) and selects the final tool.
\end{enumerate}

\begin{figure}[!t]
    \vspace*{1em}
    \centering
    \fbox{%
        \begin{minipage}{0.80\columnwidth}
            \centering
            \includegraphics[width=\linewidth]{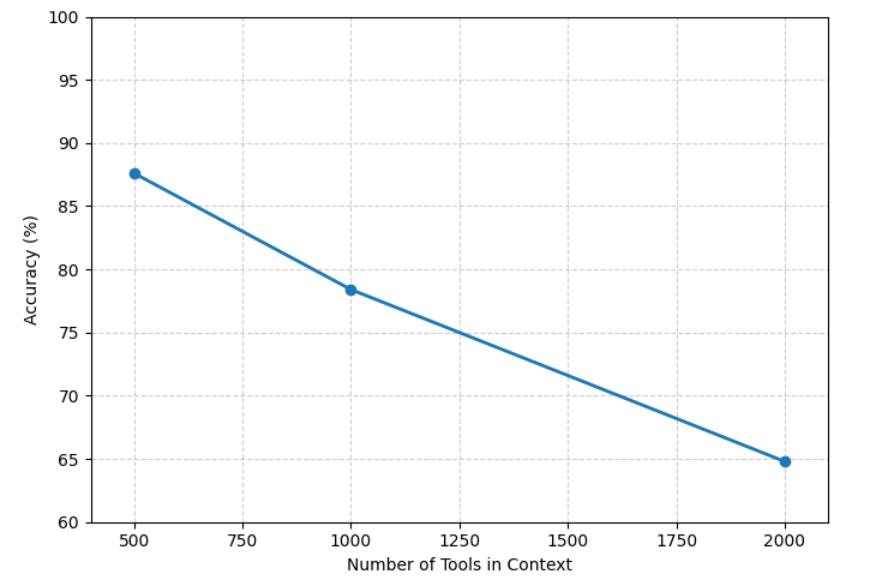}
        \end{minipage}%
    }
    \caption{Accuracy vs Tools in Context (Gemini 2.0 Flash)}
    \label{fig:long_context}
\end{figure}
\vspace{-0.7em}

\textbf{Metric:} We measured the End-to-End Top-1 Accuracy, 
reflecting the combined success of the retrieval filtering and the 
subsequent LLM reasoning.

\section{Results and Analysis}

\subsection{Cross-Model Baseline}

\begin{table}[H]
\centering
\caption{Tool Retrieval Experiment Results -- Hit Rate (\%)}
\label{tab:tool_retrieval_combined}

\resizebox{\columnwidth}{!}{%
\begin{tabular}{lcccc}
\hline
\textbf{Model} & \textbf{10 Tools} & \textbf{50 Tools} & \textbf{100 Tools} & \textbf{Avg Degradation} \\
\hline
GPT-4o-mini      & 98.2 & 92.4 & 88.2 & -10.0 \\
Claude 3.5 Haiku & 97.8 & 91.9 & 88.6 & -9.2  \\
Gemini 2.0 Flash & 98.4 & 93.6 & 88.2 & -10.2 \\
\hline
\multicolumn{5}{c}{\textit{Total runs: 9 (3 models $\times$ 3 configurations, 500 queries each).}} \\
\hline
\end{tabular}
}
\end{table}

Table \ref{tab:tool_retrieval_combined} shows a clear inverse relationship between the number of tools provided and the model's retrieval accuracy. As the context scales from 10 to 100 tools, all models experience a performance degradation of approximately 10\%. This decline indicates that as the context window fills with more candidates, the models struggle to maintain focus on the correct tool.

\textbf{Gemini 2.0 Flash} achieved the highest accuracy with a small list (98.4\% at 10 tools) but suffered the sharpest decline (-10.2 points) as the list grew. This indicates that while it is the most precise model for short contexts, its performance degrades the fastest when the context becomes crowded.

\textbf{Claude 3.5 Haiku} proved to be the most stable. Although it started with slightly lower accuracy, it experienced the smallest drop (-9.2 points) and finished with the highest score at the 100-tool level (88.6\%). This suggests it is better at retaining accuracy as it approaches its effective context limits.

\textbf{GPT-4o-mini} followed a nearly identical trend to Gemini, dropping to 88.2\% at 100 tools. Like Gemini, it handles small contexts well but loses precision at the same rate when the tool count scales up.

The experiment was capped at 100 tools due to the context window constraints of these efficiency-optimized models. The downward trend across all three suggests that testing beyond 100 tools would likely result in a significant drop in reliability.

\subsection{Large-Context Scaling}

Fig. 2 shows how Gemini 2.0 Flash performs as the tool pool expands from 500 to 2,000 tools. Accuracy decreases from 87.4\% at 500 tools to 65\% at 2,000 tools, reflecting the increased difficulty of selecting the correct tool in high-noise contexts. The sharpest drop occurs between 1,000 and 2,000 tools, suggesting that long-context interference becomes significant once the candidate set exceeds roughly one thousand items. Despite this degradation, the model maintains above-chance accuracy even at extreme scales, indicating that the dataset preserves enough semantic separation for retrieval to remain feasible at large context sizes.

\subsection{RAG Analysis}

\begin{table}[H]
\centering
\caption{Retrieval and reranking results for classical retrievers and Gemini 2.0 Flash.}
\label{tab:rag_results}
\scriptsize
\renewcommand{\arraystretch}{1.05}
\begin{tabular}{lcc}
\hline
\textbf{Retriever Model} & \textbf{Top-10 Hit Rate (\%)} & \textbf{Gemini Accuracy (\%)} \\
\hline
BM25                                 & 61.11 & 59 \\
TF--IDF                              & 75.15 & 68 \\
SentenceTransformer                  & 87.60 & 76 \\
Gemini (2,000 tools)                 & --    & 65 \\
\hline
\multicolumn{3}{c}{\textit{Top-10 for retrieval only; baseline uses full 2000 tool-context}} \\
\hline
\end{tabular}
\end{table}

Table \ref{tab:rag_results} assesses the necessity of Retrieval-Augmented Generation (RAG) when scaling to a massive 2,000-tool environment. The baseline metric shows that Gemini 2.0 Flash, when attempting to process all 2,000 tools simultaneously within its context window, achieves an accuracy of 65\%.

\textbf{SentenceTransformer} proved to be the most effective retrieval method, achieving a Top-10 Hit Rate of 87.6\%. When Gemini reranked this filtered list, it achieved a final accuracy of 76\%. This represents an 11-percentage-point improvement over the baseline, confirming that semantic retrieval successfully reduces the search space to a manageable size for the LLM.

\textbf{TF--IDF} also outperformed the baseline, yielding a final accuracy of 68\%. While its retrieval hit rate (75.15\%) was lower than the semantic model, it remained sufficient to boost Gemini's performance slightly above the raw context-processing limit.

\textbf{BM25} was the only method to underperform the baseline. With a retrieval hit rate of only 61.11\%, the resulting candidate lists were often too noisy or incomplete. Consequently, Gemini's final accuracy dropped to 59\%, proving that low-quality retrieval actively harms performance compared to simple context scaling.

Two key trends emerge from these results. First, RAG is only beneficial when the retriever outperforms the LLM's native context-handling ability; in our experiments, TF--IDF and SentenceTransformer met this threshold, while BM25 did not. Second, retrieval quality remains the primary bottleneck. The final LLM accuracy never exceeds the retrieval Top-10 hit rate, and even when the correct tool appears in the shortlist, the model still misidentifies it in roughly 10--15\% of cases. This indicates that most failures originate from the initial retrieval step, since the LLM cannot recover if the correct tool is absent from the top 10 candidates.

\vspace{0.5em}

\vspace{0.5em}

\section{Limitations}

\textbf{Persona Coverage:} The five personas offer useful variety, but they still simplify real user behavior. Actual users blend expert and novice traits, shift tone based on context or emotion, and change how they phrase queries as they learn. Cultural conventions and domain-specific communication patterns also fall outside this fixed taxonomy.

\textbf{Synthetic Query Bias:} Because every query is generated by models, the dataset reflects cleaner, more uniform language than real interactions. It underrepresents typos, fragments, code-switching, and unconventional phrasing, and it inherits stylistic assumptions from the GPT-4 family that may not generalize across user groups or future systems.

\textbf{Metadata Constraints:} Query realism depends on the clarity of each tool’s description. Tools with sparse, outdated, or overly technical metadata lead to less representative queries and cannot capture usage patterns that differ from what is written.

\textbf{Language and Cultural Limits:} The dataset is English-only, so it does not reflect multilingual query structures, culturally specific phrasing, or the diverse ways non-English speakers discover and request tools.

\textbf{Noise Design Choice:} We did not add artificial disfluency. Although this omits some messy human behavior, many deployed RAG pipelines already rewrite or clean queries upstream, so the dataset aligns with what retrievers typically receive in practice.

\section{Conclusion}

This paper introduced HumanMCP, a large-scale, persona-driven dataset designed to bridge the gap between synthetic benchmarks and realistic user behavior within the Model Context Protocol ecosystem. By leveraging a two-stage Generator-Critic pipeline, we synthesized diverse queries ranging from ambiguous problem statements to precise technical commands for 2,800 tools. 

Our evaluations demonstrate that while efficiency-optimized models perform exceptionally well in low-context settings, they suffer consistent degradation as the tool search space expands. Furthermore, our analysis confirms that in large-scale environments (2,000+ tools), distinct semantic retrieval significantly outperforms keyword-based methods and raw long-context processing. HumanMCP provides the research community with a critical resource to benchmark these retrieval challenges, enabling the development of language agents that are robust, scalable, and truly adaptive to natural human communication.

\section{ACKNOWLEDGMENT}

The authors thank the Algoverse team for their support throughout this project. Their technical guidance and constructive feedback were essential in helping shape the dataset, pipeline design, and experimental evaluations presented in this work. All authors made equivalent contributions to the research and preparation of this work.

\clearpage


\begin{thebibliography}{99}
\bibitem{hou2025mcp} X.~Hou, Y.~Zhao, S.~Wang, and H.~Wang, `Model Context Protocol (MCP): Landscape, Security Threats, and Future Research Directions,'' \emph{arXiv preprint arXiv:2503.23278}, 2025.

\bibitem{mcp_tools_docs} Model Context Protocol Project, `Tools -- Model Context Protocol (MCP) Documentation,'' 2025. [Online]. Available: \url{https://modelcontextprotocol.io/docs/concepts/tools}


\bibitem{brown2020language} T.~B.~Brown \emph{et al.}, `Language Models are Few-Shot Learners,'' in \emph{Advances in Neural Information Processing Systems}, 2020.

\bibitem{openai2023gpt4} OpenAI, `GPT-4 Technical Report,'' \emph{arXiv preprint arXiv:2303.08774}, 2023.

\bibitem{metagpt} H.~Zhuge \emph{et al.}, `MetaGPT: Meta Programming for a Multi-Agent Collaborative Framework,'' in Proc. ICLR, 2024. [Online]. Available: \url{https://arxiv.org/abs/2308.00352v3}

\bibitem{autogen} Microsoft Research, `AutoGen: An open-source framework for agentic AI,'' 2024. [Online]. Available: \url{https://www.microsoft.com/en-us/research/project/autogen/}

\bibitem{mcp_spec} Model Context Protocol Project, `Specification -- Model Context Protocol (MCP),'' 2025. [Online]. Available: \url{https://modelcontextprotocol.io/specification/2025-03-26}

\bibitem{mcp-agentbench} Z.~Guo \emph{et al.}, `MCP-AgentBench: Evaluating Real-World Language Agent Performance with MCP-Mediated Tools,'' \emph{arXiv preprint arXiv:2509.09734}, 2025.

\bibitem{gao2025mcp-radar} X.~Gao, S.~Xie, J.~Zhai, S.~Ma, and C.~Shen, `MCP-RADAR: A Multi-Dimensional Benchmark for Evaluating Tool Use Capabilities in Large Language Models,'' \emph{arXiv preprint arXiv:2505.16700}, 2025.

\bibitem{fan2025mcptoolbenchpp} Z.~Fan \emph{et al.}, `MCPToolBench++: A Large Scale AI Agent Model Context Protocol MCP Tool Use Benchmark,'' \emph{arXiv preprint arXiv:2508.07575}, 2025.

\bibitem{wang2025mcpbench} Y.~Wang \emph{et al.}, `MCP-Bench: Benchmarking Tool-Using LLM Agents with Complex Real-World Tasks via MCP Servers,'' \emph{arXiv preprint arXiv:2508.20453}, 2025.

\bibitem{guo2025mcpagentbench} Z.~Guo \emph{et al.}, `MCP-AgentBench: Evaluating Real-World Language Agent Performance with MCP-Mediated Tools,'' \emph{arXiv preprint arXiv:2509.09734}, 2025.

\bibitem{xu2025toucan} Z.~Xu \emph{et al.}, `TOUCAN: Synthesizing 1.5M Tool-Agentic Data from Real-World MCP Environments,'' \emph{arXiv preprint arXiv:2510.01179}, 2025.

\bibitem{comanici2025gemini} G.~Comanici \emph{et al.}, `Gemini 2.5: Pushing the frontier with advanced reasoning, multimodality, long context, and next generation agentic capabilities,'' \emph{arXiv preprint arXiv:2507.06261}, 2025.

\bibitem{achiam2023gpt} J.~Achiam \emph{et al.}, `GPT-4 Technical Report,'' \emph{arXiv preprint arXiv:2303.08774}, 2023.

\bibitem{rahman2024automatic} M.~Rahman, S.~Khatoonabadi, A.~Abdellatif, and E.~Shihab, `Automatic detection of LLM-generated code: A case study of Claude 3 Haiku,'' \emph{arXiv preprint arXiv:2409.01382}, 2024.

\bibitem{alsini2020hit} A.~Alsini, D.~Q.~Huynh, and A.~Datta, `Hit ratio: An evaluation metric for hashtag recommendation,'' \emph{arXiv preprint arXiv:2010.01258}, 2020.

\bibitem{fei2025mcp} X.~Fei, X.~Zheng, and H.~Feng, `MCP-Zero: Proactive Toolchain Construction for LLM Agents from Scratch,'' \emph{arXiv preprint arXiv:2506.01056}, 2025.

\bibitem{bafna2016document} P.~Bafna, D.~Pramod, and A.~Vaidya, `Document clustering: TF-IDF approach,'' in \emph{Proc. Int. Conf. Electrical, Electronics, and Optimization Techniques (ICEEOT)}, 2016, pp.~61--66.

\bibitem{robertson1994simple} S.~Robertson and K.~Sp\"{a}rck Jones, `Simple, proven approaches to text retrieval,'' Tech. Rep., Univ. of Cambridge, 1994.

\bibitem{reimers2019sentence} N.~Reimers and I.~Gurevych, `Sentence-BERT: Sentence embeddings using siamese BERT-networks,'' \emph{arXiv preprint arXiv:1908.10084}, 2019.

\end{thebibliography}
\end{document}